\pdfoutput=1

\documentclass[11pt]{article}

\usepackage[final]{acl}

\usepackage{times}
\usepackage{latexsym}

\usepackage[T1]{fontenc}

\usepackage[utf8]{inputenc}

\usepackage{microtype}

\usepackage{inconsolata}

\usepackage{graphicx}
\usepackage{soul}
\usepackage{xcolor,color}
\usepackage{multirow}
\usepackage{float}
\usepackage{array}
\usepackage{booktabs}
\usepackage{amsmath}
\usepackage{amssymb}
\usepackage{makecell}
\usepackage{hyperref}
\usepackage{setspace}
\usepackage{enumitem}

\definecolor{lightblue}{RGB}{205,230,247}
\definecolor{lightgreen}{RGB}{213,242,202}
\definecolor{lightyellow}{RGB}{255,255,170}

\title{Enhance Multimodal Consistency and Coherence \\ for Text-Image Plan Generation}

\author{Xiaoxin Lu~~~~Ranran Haoran Zhang~~~~Yusen Zhang~~~~Rui Zhang\\
  The Pennsylvania State University, State College, PA, USA \\
  \texttt{\{xzl5514, haoranz6, yfz5488, rmz5227\}@psu.edu}}

\begin{document}
\maketitle

\begin{abstract}

People get informed of a daily task plan through diverse media involving both texts and images. 
However, most prior research only focuses on LLM's capability of textual plan generation. The potential of large-scale models in providing text-image plans remains understudied. Generating high-quality text-image plans faces two main challenges: ensuring consistent alignment between two modalities and keeping coherence among visual steps.
To address these challenges, we propose a novel framework that generates and refines text-image plans step-by-step. 
At each iteration, our framework (1) drafts the next textual step based on the prediction history; (2) edits the last visual step to obtain the next one; (3) extracts PDDL-like visual information; and (4) refines the draft with the extracted visual information. The textual and visual step produced in stage (4) and (2) will then serve as inputs for the next iteration. Our approach offers a plug-and-play improvement to various backbone models, such as Mistral-7B, Gemini-1.5, and GPT-4o. To evaluate the effectiveness of our approach, we collect a new benchmark consisting of 1,100
tasks and their text-image pair solutions covering 11
daily topics. We also design and validate a new set of metrics to evaluate the multimodal consistency and coherence in text-image plans. Extensive experiment results show the effectiveness of our approach on a range of backbone models against competitive baselines. Our code and data are available at \url{https://github.com/psunlpgroup/MPlanner}.

\end{abstract}

\section{Introduction}

\begin{figure}[t!]
\centering
\vspace{-5mm}
\includegraphics[width=0.5\textwidth]{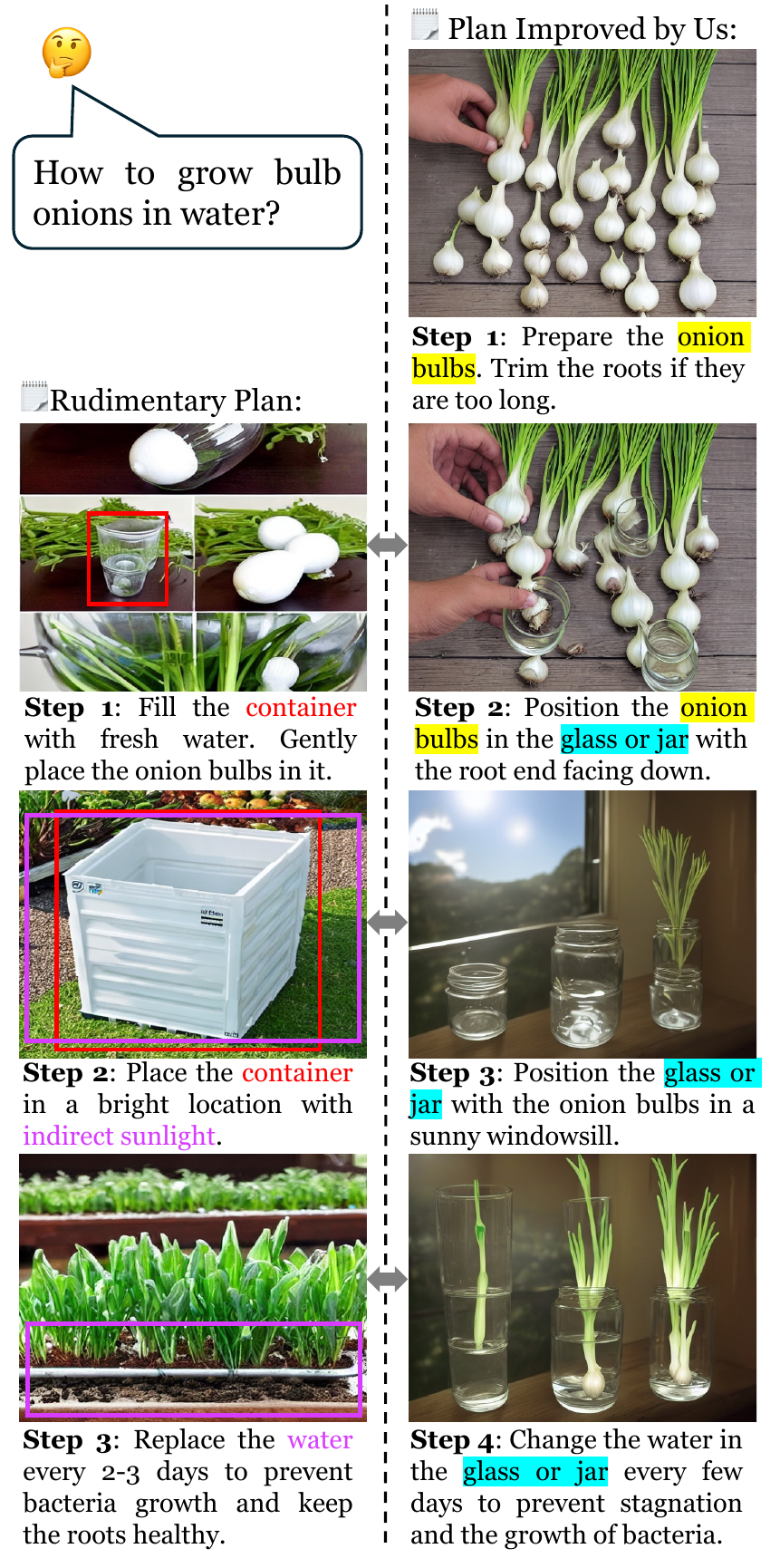}
\caption{
Plans generated by GPT-4o (left)
and our framework (right).
Our framework maintains higher consistency between images and texts, and achieves higher coherence among images across different steps.
}
\vspace{-3mm}
\label{fig:intro-case}
\end{figure}

Recently, there has been growing attention on employing LLMs for planning, the task of decomposing a high-level goal into a sequence of executable steps~\cite{valmeekam2022large,hao2023reasoning}. 
LLMs have demonstrated strong capabilities in generating textual plans, enabling applications in robotics, virtual assistants, and instructional content generation~\citep{huang2022languagemodelszeroshotplanners,liu2023llmpempoweringlargelanguage,silver2024generalized}.
However, textual plans alone can be insufficient, as many real-world tasks require both textual instructions and visual demonstrations for clarity. Multimodal task planning, which can be formulated as a paired text-image sequence generation problem conditioned on the task goal~\cite{lu-etal-2024-multimodal}, enhances comprehensibility and usability by leveraging the complementary strengths of language and vision. Despite its potential, multimodal planning remains an underexplored problem, with prior approaches struggling to maintain consistency between modalities and coherence across visual steps.

Multimodal task planning faces two main challenges: ensuring coherence through visual steps and alignment between two modalities. Figure~\ref{fig:intro-case} illustrates these challenges through a comparative example of generating plans for the input goal ``growing bulb onions in water''. The left-hand side shows the plan generated by GPT-4o,
and the right-hand side presents the corresponding plan generated by our framework.
The baseline plan, although generally reasonable, exhibits critical issues in both visual coherence and text-image alignment. First, it fails to maintain visual consistency between step 1 and step 2, as highlighted in red. The container depicted in step 1 is a transparent glass, but in step 2, it becomes an opaque white planter, disrupting coherence. Second, as highlighted in purple, step 2's image shows a location with direct sunlight, contradicting the textual instruction that specifies ``indirect sunlight''. Furthermore, step 3 entirely diverges from the intended goal of growing onions \textit{in water}, as it depicts bulbs planted in soil.

On the other hand, our framework effectively improves visual coherence and text-image consistency. As shown on the right-hand side of Figure~\ref{fig:intro-case}, our approach maintains a uniform depiction of the onion bulbs and their container throughout the process. The glass or jar remains consistent in shape and size from step 2 to step 4, avoiding abrupt visual changes. Moreover, the visual steps accurately reflect the textual descriptions through all steps. For instance, step 3 correctly depicts a sunny windowsill, aligning with the text's instruction to position the jar in such a location.

These improvements stem from our novel autoregressive framework. At each iteration, the text generator drafts the current step based on the task goal and previous steps. The image generator then produces a corresponding visual representation conditioned on the last visual step. An image interpreter subsequently extracts structured information from the generated image, which the text generator uses to refine the step draft. This iterative process enhances visual coherence with a text-image-to-image model and ensures text-image consistency through cross-modality prompting. Compared to vanilla approaches that simply concatenate an LLM with a text-to-image model, our framework effectively mitigates the challenges of multimodal task planning, leading to more coherent and consistent instructional sequences.

To evaluate our framework, we collect a dataset from two popular websites: Instructables and wikiHow. Our dataset consisting of 1100 examples in 11 categories,
providing rich multimodal information of procedural solutions to a variety of daily tasks. We adopt a set of metrics including conventional automatic measurements, LLM evaluations, and human evaluations to comprehensively evaluate planning performance in three aspects: 
textual plan quality, visual plan quality, and textual-visual plan alignment. 
To demonstrate the generalizability of our approach, we evaluate our framework with 3 different backbones: Mistral-7B~\cite{jiang2023mistral}, Gemini-1.5-flash~\cite{team2024gemini}, and GPT-4o~\cite{hurst2024gpt}. For every backbone, we compare our framework with various baselines. Extensive experiment results show our framework outperforms all baselines, especially in terms of the two concerns we aim to address. 

In summary, our contributions are three-fold:
\begin{itemize}[noitemsep,topsep=0pt]
    \item We propose a novel framework to address both visual coherence challenge and text-image alignment challenge of multimodal planning problem; 
    \item We collect a dataset of daily tasks covering diverse domains and complexity levels to evaluate the text-image planning performance. 
    \item We empirically show the effectiveness of our framework with extensive experimental results and visualization examples.
\end{itemize}










\vspace{-2mm}
\section{Related Work}
\vspace{-2mm}
\paragraph{Task Planning}

Task planning is broadly studied in various scenarios of virtual environment \citep{zhao2023largelanguagemodelscommonsense,gao2023largelanguagemodelsempowered,hu2024chainofsymbolpromptingelicitsplanning}, embodied environment \citep{huang2022languagemodelszeroshotplanners,song2023llmplannerfewshotgroundedplanning,zhang2024instructlargelanguagemodels}, and daily life \citep{oswald2024largelanguagemodelsplanning,wu-etal-2022-understanding}. Despite classical planning algorithms \citep{Hoffmann_2001,alarnaouti2023reformulationtechniquesautomatedplanning}, they are primarily applicable with restrictions such as fully observable environments with pre-defined actions and objects.
It prohibits their usage in open-domain daily scenarios, igniting research interest in solutions from LLM advancements \citep{kambhampati2024llmscantplanhelp}. 

LLMs, possessing a rich amount of commonsense knowledge and impressive reasoning capability, are competent in such contexts. \citet{huang2022languagemodelszeroshotplanners} studies LLMs as zero-shot planners and shows their great planning potential in a virtual environment. \citet{liu2023llmpempoweringlargelanguage} combines the LLM with classical planners to get reliable solutions to robotic tasks in an embodied environment. \citet{arora2023learningleveragingverifiersimprove} tries to add an external verifier to improve the planning capabilities of a finetuned LLM. \citet{wang-etal-2023-multimedia} seeks to study textual plan generation provided visual states as supplements. Despite these, only rare efforts contribute to studying multimodal planning. \citet{lu-etal-2024-multimodal} first explores the potential of LLMs and text-to-image models to generate image-text paired plans. However, their approach is limited by the single-shot generation framework and reliance on image captions that inadequately capture object interactions, leading to information loss and visual incoherence in planning sequences.

\vspace{-2.5mm}
\paragraph{Vision and Language}

Diffusion models \citep{ho2020denoisingdiffusionprobabilisticmodels,ramesh2021zeroshottexttoimagegeneration,rombach2022highresolutionimagesynthesislatent} have revolutionized image generation conditioned on texts. 
However, in specific circumstances where the models are expected to output coherent images, they exhibit poor performance because of the lack of visual context knowledge \citep{lu-etal-2024-multimodal}.
This leads to the introduction of advanced image editing models \citep{kawar2023imagictextbasedrealimage}. Building upon the success of image generation models, they are capable of generating images conditioned on both reference images and text instructions. For example, \citet{brooks2023instructpix2pixlearningfollowimage} achieves image editing by fine-tuning the stable diffusion model on (original image, instruction, edited image) triplets. \citet{zhang2023addingconditionalcontroltexttoimage} enables precise spatial conditioning controls including edges, depth maps, pose information, etc. \citet{Soucek_2024_CVPR} infuses video data into a diffusion model to generate images depicting how actions lead to object state transformations. However, they all tend to maintain the outline of objects in the original image while only editing the color, texture, or style. Thus, they are still inadequate in generating coherent images for planning, which typically involves scenario change and object transformation.
\vspace{-3.5mm}
\section{Dataset}
\vspace{-2mm}
\label{section-dataset}
Despite existing datasets in the daily task planning area, they either lack image modality \citep{koupaee2018wikihowlargescaletext,valmeekam2023planbench}, task domain diversity \citep{yagcioglu2018recipeqachallengedatasetmultimodal,valmeekam2023planning}, or are not intended for plan generation \citep{yang2021visualgoalstepinferenceusing}. Therefore, to benchmark text-image plan generation, we collect a dataset of daily task plans covering various topics.

\begin{figure}[t!]
    \centering
    \includegraphics[width=0.48\textwidth]{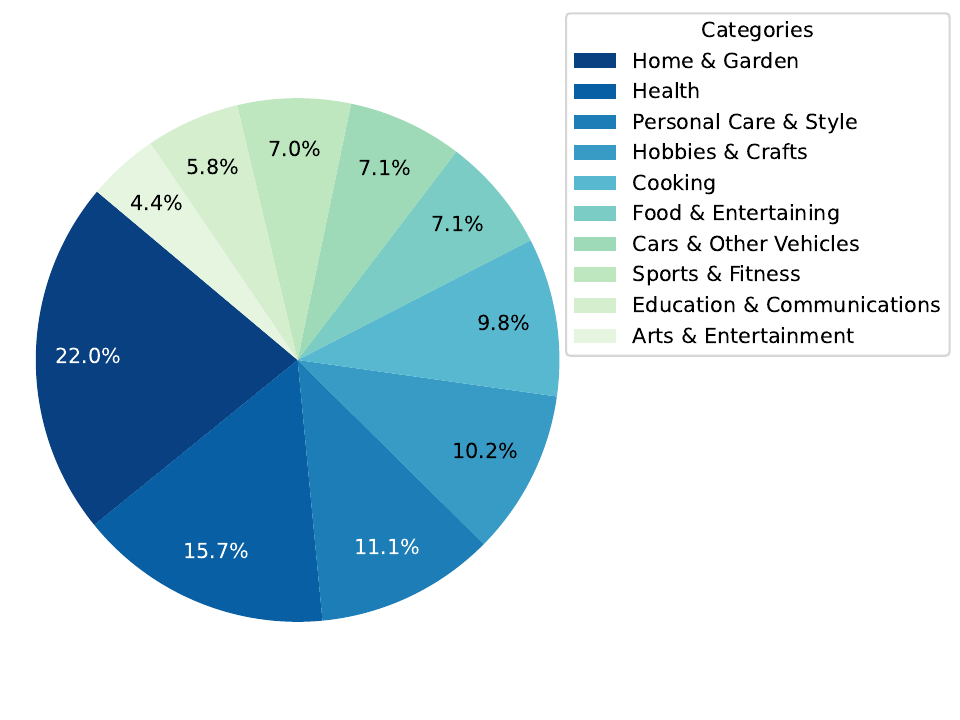}
    \vspace{-12mm}
    \caption{
    The distribution of our dataset.
    }
     \vspace{-3mm}
    \label{fig:data-distribution}
\end{figure}

Inspired by previous work, we consider two popular websites affording procedural daily task instructions: Instructables\footnote{\url{https://www.instructables.com/}} and wikiHow\footnote{\url{https://www.wikihow.com/}}. Both data sources provide various modalities and cover diverse daily task categories.
The overall category distribution of our dataset is shown in Figure \ref{fig:data-distribution}.  Compared with previous public datasets, our dataset is rich in task categories and plan modalities. Furthermore, we ensure high data quality by manually filtering out malicious content and curating the collected plans to remove noises such as the authors' information and personal stories. Please see Appendix~\ref{sec:app-dataset} for more dataset statistics and demonstrations.

\begin{figure*}[t!]
    \centering
    \includegraphics[width=1.0\textwidth]{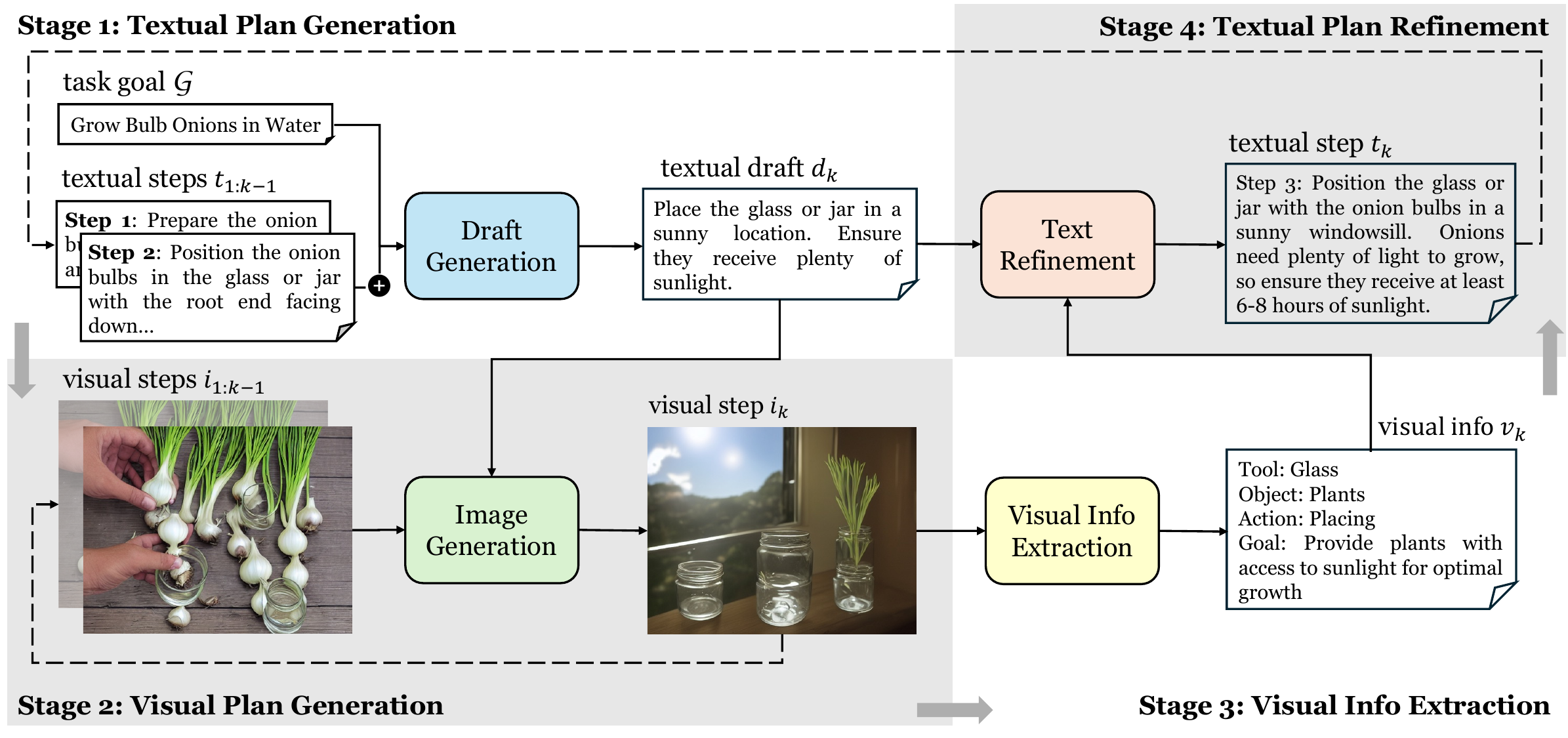}
    \caption{Overview of our autoregressive framework at time step $k$. Stage 1: the text plan generator takes the input task goal $\mathcal{G}$ and output textual steps $t_{1:k-1}$ from previous time steps to predict $d_k$. Stage 2: the image generator outputs the visual step $i_k$ conditioned on $d_k$ and the last visual step $i_{k-1}$. Stage 3: the text generator extracts formatted visual information from the generated $i_k$ as $v_k$. Stage 4: the text generator refines $d_k$ with $v_k$ to generate $t_k$. Dotted Arrow: The output $t_k$ and $i_k$ then serve as the input for the next time step $k+1$.}
    \label{fig:framework}
\end{figure*}
\vspace{-2mm}
\paragraph{Instructables}

To study how our framework performs on simple daily tasks, we randomly collect 100 plans from Instructable. Specifically, we choose ``Cooking'' category as it is a common planning scenario. Plans from Instructables are published by users who would like to share their experience in achieving a specific goal. Therefore, their plans are more brief and more casual.
\vspace{-2mm}
\paragraph{wikiHow}

We randomly sample 1,000 plans from wikiHow expert articles. In 19 categories, we select 10 that best fit into the multimodal planning context. The remaining categories, such as family life, relationships, philosophy and religion, etc, often involve sensitive information. Moreover, they are not suitable for our planning problem since the provided ``plans'' are more like ``advice'' without explicit temporal consistency and dependency through the ``steps''. Our collected wikiHow articles are reviewed, edited, or authored by domain experts verified by the platform. Therefore, these plans are longer, more detailed, and more formal.

\vspace{-2mm}
\section{Approach}
\vspace{-2mm}

\paragraph{Task Formulation}

Given a task goal $\mathcal{G}$ as the input, the output of multimodal planning is a sequence of steps $\mathcal{S}=\{s_1,s_2,...,s_n\}$. Each step consists of an instructional text $t_k$ and its corresponding image $i_k$, so we have $s_k=(t_k,i_k), k\in\{1,2,...,n\}$. At every step $k$, $t_k$ and $i_k$ should be semantically consistent. Both textual plan $\mathcal{S}_t=\{t_1,...,t_n\}$ and visual plan $\mathcal{S}_i=\{i_1,...,i_n\}$ should be coherent through steps.
\vspace{-2mm}
\paragraph{Framework Overview}
As shown in Figure \ref{fig:framework}, our framework generates plan steps iteratively. At each step, 4 cross-modality stages collaboratively contribute to the multimodal plan generation. In the first stage, the textual plan drafter takes the input task goal $\mathcal{G}$ and the previous textual steps $t_{1:k-1}$ to draft the current textual step $d_k$. In stage two, the image generator edits the last visual step $i_{k-1}$ conditioned on $d_k$ to get the current visual step $i_k$. In stage three, an image interpreter extracts formatted visual information $v_k$ from the generated $i_k$. Last, the textual plan refiner collects $v_k$ and uses it to refine the original textual step draft $d_k$ to be $t_k$. After the current iteration is complete, both $t_k$ and $i_k$ are added to history steps for the next iteration. The overall design ensures the coherence of visual steps by prompting the image generation module to predict the next image based on both textual instruction and the previous visual state. Combining text-to-image and converse image-to-text in a loop guarantees consistency between the generated textual plan and its visual counterpart. Detailed prompts for each module are in Appendix~\ref{app-exp-settings}.
\vspace{-2mm}
\subsection{Textual Plan Drafting}

As LLMs embed a rich amount of commonsense knowledge, we prompt them with the task goal $\mathcal{G}$ at the first iteration to get the textual step draft $d_1$. At later iterations, we concatenate $\mathcal{G}$ with previous textual steps $t_{1:k-1}$ to draft $d_k$:

\begin{small}
\begin{equation}d_k=
    \begin{cases}
        \mathbf{G}_t(\mathcal{G}),&k=1\\
        \mathbf{G}_t(\mathcal{G}, \text{Concat}(t_1, t_2, ..., t_{k-1})),&k>1
    \end{cases}
\end{equation}
\end{small}
\noindent where $\mathbf{G}_t$ denotes the text generation model.

\subsection{Visual Plan Generation}

To maintain coherence through visual steps, we employ InstructPix2Pix \citep{brooks2023instructpix2pixlearningfollowimage}, a prevailing image editing model. Different from image generation models, InstructPix2Pix is conditioned on both the text prompt and the input image. In our framework, it ensures the generation of the current visual step $i_k$ is always aware of the last visual state $i_{k-1}$. This helps improve the coherence through the visual plan $\mathcal{S}_i=\{i_1,i_2,...,i_n\}$.

\begin{equation}
    i_k=\mathbf{G}_i(d_k, i_{k-1})
\end{equation}
\noindent where $\mathbf{G}_i$ denotes the image generation model.



\subsection{Textual Plan Refinement}
We further employ a textual plan refinement mechanism for two reasons: 1) to improve the consistency between textual step and visual step; 2) to complement the textual step draft with implicit knowledge embedded in the image.

\paragraph{Visual Information Extraction}
First, we adopt a visual information-infused model as the image interpreter to extract visual information. Specifically, we adopt the idea of planning domain definition language (PDDL) from the classical planning field \citep{Fox_2003}. We design a pseudo-PDDL (pPDDL) with structured representations that are applicable in our planning scenario. As displayed in Figure \ref{fig:framework}, we specify this pPDDL as 4 different types of information embedded in a single image: involved objects, tools, the action, and the goal. 
\begin{equation}
    v_k=\mathbf{E}(i_k)
\end{equation}
\noindent $\mathbf{E}$ is a language model to generate pPDDL.

\vspace{-2mm}
\paragraph{Visual Information Incorporation}
Last, we feed the extracted visual information into the plan generator to revise the textual step draft. 
\begin{equation}
    t_k=\mathbf{G}_t(d_k, v_k)
\end{equation}
\section{Experiment Settings}



\vspace{-2mm}
\subsection{Baselines}
\vspace{-2mm}
To demonstrate the effectiveness of our model, we compare with three types of strong multimodal planning baselines based on state-of-the-art LLMs and diffusion models:
\begin{itemize}[noitemsep,topsep=0pt]
    \item Vanilla LLM baselines including \textsc{Gemini-1.5-Flash}, \textsc{Gpt-4o}, and LLaVa with instruction finetuning using Mistral (\textsc{M\&L}). The backbone LLM generates textual plan in one turn, and Stable Diffusion model generates the visual plan according to parsed textual steps. 
    \item \textsc{Sd}: Stable Diffusion~\citep{rombach2022highresolutionimagesynthesislatent} generates one visual step at each turn given the task goal $\mathcal{G}$, and the backbone LLM describes each visual step to form a textual plan. 
    \item We also compare with a state-of-the-art multimodal planning framework \textsc{Tip}~\citep{lu-etal-2024-multimodal}. Compared with the first type baseline, it leverages cross-modality prompting with a T2I-Bridge and an I2T-Bridge to further improve the performance. For fair comparison, we use \textsc{Tip} on top of the same backbone language model choices as ours.
\end{itemize}

\subsection{Evaluation Metrics}

\paragraph{Textual Plan Evaluation}
We conventionally choose BertScore \citep{zhang2020bertscoreevaluatingtextgeneration} and ROUGE (R-1, R-2, R-L) \citep{lin-2004-rouge} to automatically measure the semantic similarity between generated textual plans and reference textual plans. Considering the open-ended attribute of the daily task planning problem, automatic metrics do not suffice to evaluate plan quality. Therefore, we also include four qualitative metrics: correctness, executability, coherence, and informativeness. We use Claude-3.5-Sonnet as the LLM judge to score generated plans with awareness of reference plans and have three human annotators verify the LLM evaluation's reliance. Please see Appendix~\ref{app-exp-settings} for more details.

\paragraph{Text-Image Evaluation}
Following the common practice, we use CLIP score to examine the semantic alignment between textual steps and their visual counterparts. Like textual plan evaluation and visual plan evaluation, we use an LLM judge and human annotators to perform evaluations as well.

\paragraph{Visual Plan Evaluation}
Evaluating visual plans is nontrivial due to four cases through the plan: scenario coherence/change when actions are conducted in the same workplace/different workplaces; object coherence/change when involved objects are unchanged/changed in their states by some actions. Therefore, conventional image similarity measurements like FID~\cite{heusel2017gans} do not fit into this context. To this end, we first convert every image to a textual description of the background, salient objects, and the involved action (if any). Then, we employ the perplexity score (PPL) to check if consecutive descriptions are coherent \textit{given the action corresponding to the later step}. In addition, we also conduct LLM evaluations and human evaluations.

\begin{table*}[ht]
\small
\begin{center}
\renewcommand{\arraystretch}{1.2}
\resizebox{\textwidth}{!}{
\begin{tabular}{clcccccccccccc}
\toprule
\multirow{2}{*}{\makecell{Backbone\\ \& \\Dataset}}       & \multirow{2}{*}{\raisebox{-.5\height}{Method}} & \multicolumn{6}{c}{Automatic Evaluation}  & \multicolumn{6}{c}{LLM Evaluation}                        \\ \cmidrule(lr){3-8} \cmidrule(lr){9-14} 
                               &                         & BertScore $\uparrow$ & R-1 $\uparrow$ & R-2 $\uparrow$ & R-L $\uparrow$ & CLIP $\uparrow$ & PPL $\downarrow$ & Corr. $\uparrow$ & Exec. $\uparrow$ & Coh. $\uparrow$ & Info. $\uparrow$ & T-I $\uparrow$ & I-I $\uparrow$ \\ \midrule
\multirow{4}{*}{\makecell{Gemini\\ \& \\Instructables}} & \textsc{Gemini}                  & \underline{0.835}     & \underline{26.3}   & \underline{7.00}   &                                 \underline{24.7}   & 17.84        & 5.98     & \underline{4.81}  & \underline{4.84}  & \textbf{4.96}  & \underline{4.85}                                   & 1.59 & \underline{2.33}\\
                               & \textsc{Sd}                     & 0.807     & 20.7   & 5.20   & 19.7   & 8.59        & \textbf{4.85}   & 1.17  & 0.74  & 0.85  & 2.33  & 1.04  & 1.29 \\
                               & \textsc{Tip}                     & 0.812     & 22.2   & 5.00   & 20.5   & \underline{18.66}        & 6.11    & 3.15  & 3.10  & 3.54  & 3.57  & \underline{1.73}  & 2.15\\
                               & \textsc{Ours}                    & \textbf{0.842}     & \textbf{30.6}   & \textbf{9.20}   & \textbf{28.2}   & \textbf{26.38}        & \underline{5.49}    & \textbf{4.85}  & \textbf{4.89}  & \underline{4.91}  & \textbf{4.94}  & \textbf{2.64}  & \textbf{2.41}\\
                               \midrule
\multirow{4}{*}{\makecell{Gemini\\ \& \\wikiHow}}       & \textsc{Gemini}                  & \underline{0.847}     & \underline{26.1}   & \underline{7.30}   & \underline{24.4}   & \underline{15.61}        & 5.92     & \underline{4.83}  & \underline{4.84}  & \textbf{4.97}  & \underline{4.82}  &                                \underline{1.69}  & \underline{2.32}\\
                               & \textsc{Sd}                    & 0.806     & 14.4   & 2.00   & 13.7   & 9.03        & \textbf{4.94}    & 0.79  & 0.70  & 0.81  & 1.94  & 1.28  & 1.13 \\
                               & \textsc{Tip}                     & 0.812     & 21.5   & 5.40   & 20.4   & 14.58        & 6.01     & 3.32  & 3.47  & 3.51  & 3.82  & 1.58  & 2.31 \\
                               & \textsc{Ours}                    & \textbf{0.850}     & \textbf{29.0}   & \textbf{9.00}   & \textbf{27.0}   & \textbf{20.23}        & \underline{5.13}    & \textbf{4.89}  & \textbf{4.88}  & \underline{4.89}  & \textbf{4.91}  & \textbf{2.42}  & \textbf{2.40} \\
                               \bottomrule
                               \toprule

\multirow{4}{*}{\makecell{GPT\\ \& \\Instructables}} & \textsc{Gpt}                  & 0.827      & 27.8   & 7.40                                   & 26.0   & 12.32    & 5.75      & \underline{4.90}       & \underline{4.87}      & \textbf{4.97}      & \underline{4.83}      & 1.53      & \underline{2.47}\\
                               & \textsc{Sd}                     & 0.805      & 19.4   & 4.30   & 18.4   & 9.65        & \textbf{5.09}      & 1.33       & 0.92      & 0.79      & 2.04      & 1.10      & 1.24\\
                               & \textsc{Tip}                     & \underline{0.840}      & \underline{29.8}   & \underline{8.20}   & \underline{28.0}   & \underline{13.19}          & 6.27      & 3.78       & 3.25      & 3.63      & 3.61      & \underline{1.68}      & 2.30\\
                               & \textsc{Ours}                    & \textbf{0.849}      & \textbf{33.7}   & \textbf{10.3}   & \textbf{31.5}   & \textbf{27.14}          &  \underline{5.21}       & \textbf{4.93}      & \textbf{4.90}       & \underline{4.93}      & \textbf{4.93}      & \textbf{2.47}      & \textbf{2.76}\\
                               \midrule
\multirow{4}{*}{\makecell{GPT\\ \& \\wikiHow}}       & \textsc{Gpt}                  & \underline{0.850}      & \underline{30.0}   & \underline{9.00}   &                                   \underline{28.1}    & 11.29    & 5.83      & \underline{4.84}       & \underline{4.86}      & \textbf{4.97}      & \underline{4.87}      & 1.56      & \underline{2.35}\\
                               & \textsc{Sd}                     & 0.811      & 20.3   & 4.80   & 19.2    & 10.37    & \textbf{5.17}       & 1.19      & 0.80      & 0.84      & 2.12      & 1.08      & 1.13\\
                               & \textsc{Tip}                     & 0.843      & 29.9   & 8.70   & 27.9    & \underline{11.81}   & 5.97      & 3.50       & 3.71      & 3.88      & 3.68      & \underline{1.73}      & 2.21\\
                               & \textsc{Ours}                    & \textbf{0.856}      & \textbf{33.2}   & \textbf{10.5}   & \textbf{30.9}    & \textbf{24.62}  & \underline{5.30}       & \textbf{4.88}      & \textbf{4.91}      & \underline{4.90}      & \textbf{4.94}      & \textbf{2.58}      & \textbf{2.68}\\
                               \bottomrule
                               \toprule
\multirow{4}{*}{\makecell{M\&L\\ \& \\Instructables}} & \textsc{M\&L}                  & 0.829      & \underline{30.8}   & 9.10   &                                  28.8    & 19.36    & 6.03     & \underline{4.79}     & \underline{4.58}     & \textbf{4.80}     & \underline{4.67}     & 1.51     & \underline{2.26}\\
                               & \textsc{Sd}                     & 0.807     & 20.7   & 5.20   & 19.7    & 11.21     & \textbf{5.31}      & 0.83     & 0.75     & 0.71     & 1.96     & 1.06     & 1.25\\
                               & \textsc{Tip}                     & \underline{0.842}      & 30.5   & \underline{9.50}   & \underline{29.3}    & \underline{20.07}   & 5.99      & 4.02     & 3.85     & 4.07     & 3.73     & \underline{1.82}     & 2.18\\
                               & \textsc{Ours}                    & \textbf{0.848}      & \textbf{32.5}   & \textbf{10.1}   & \textbf{30.1}    & \textbf{26.39}  & \underline{5.47}      & \textbf{4.81}     & \textbf{4.74}     & \underline{4.66}     & \textbf{4.74}     & \textbf{2.58}     & \textbf{2.70}\\
                               \midrule
\multirow{4}{*}{\makecell{M\&L\\ \& \\wikiHow}}       & \textsc{M\&L}                  & 0.836      & 30.0   & \underline{9.00}   &                                  \underline{28.1}    & \underline{15.82}    & 6.19      & \underline{4.75}     & \underline{4.62}     & \textbf{4.77}     & \underline{4.60}     & 1.53     & \underline{2.29}\\
                               & \textsc{Sd}                     & 0.809     & 20.0   & 4.40   & 19.3    & 10.19    & \textbf{5.20}      & 0.78     & 0.77     & 0.80     & 2.03     & 1.10     & 1.19\\
                               & \textsc{Tip}                     & \underline{0.839}      & \underline{30.1}   & 8.90   & 27.8    & 15.47    & 6.16      & 3.84     & 3.68     & 3.92     & 3.70     & \underline{1.84}     & 2.07\\
                               & \textsc{Ours}                    & \textbf{0.851}      & \textbf{31.7}   & \textbf{10.0}   & \textbf{29.6}   &\textbf{19.60}    & \underline{5.27}      & \textbf{4.83}     & \textbf{4.75}     & \underline{4.71}     & \textbf{4.78}     & \textbf{2.45}     & \textbf{2.53}\\
                               \bottomrule
\end{tabular}}
\end{center}
\vspace{-3mm}
\caption{Main results.
The best results are in \textbf{bold}, and the second best results are \underline{underlined}.
Textual Plan Evaluation: BertScore, R-1, R-2, R-L, Corr.(correctness), Exec.(executability), Coh.(coherence), Info.(informativeness).
Text-Image Evaluation: CLIP, T-I (text-image).
Visual Plan Evaluation: PPL, I-I (image-image).
}
\vspace{-3mm}
\label{table:main-results}
\end{table*}

\subsection{Implementation Details}

\paragraph{Backbone LLMs}


To examine the effectiveness of our approach on both open-source and closed-source LLMs, we choose 3 different models as the backbone: \textsc{mistral-7B}, \textsc{gemini-1.5-flash}, and \textsc{gpt-4o}. In all experiments, the backbone LLM works both as the draft generator and the text refinement editor. Since \textsc{gemini-1.5-flash} and \textsc{gpt-4o} can also interpret images, we use them as the visual information extractor in our framework. For experiments with \textsc{mistral-7B} backbone, we choose \textsc{instructblip-vicuna-7b}, a general-purpose MLLM tuned on diverse tasks, to extract visual information in the framework.
\vspace{-3mm}
\paragraph{InstructPix2Pix Finetuning}
To make the image generator better align with our context, we fine-tune \textsc{InstructPix2Pix} on a re-purposed dataset collected from wikiHow \citep{yang2021visualgoalstepinferenceusing}. The original dataset includes more than 60,000 tasks covering all categories in wikiHow. We sample 20,000 of them in the scope of our selected categories as shown in Figure \ref{fig:data-distribution}. Furthermore, we transform each plan into a series of $\{i_{k-1},t_k,i_k\}$ triplets, where $i_{k-1}$ and $i_k$ are two consecutive images, and $t_k$ the text corresponding to the later image. The fine-tuning aims to promote \textsc{InstructPix2Pix}'s capability to generate coherent $i_k$ given $i_{k-1}$ and $t_k$ as input. We split the re-purposed dataset by tasks in 0.9/0.05/0.05 for training/validation/test. We follow most training hyperparameters in the original work but change the maximum number of epochs to 50. At the end of training, we achieve training loss of 0.100 and validation loss of 0.105.
\vspace{-3mm}
\section{Results and Analysis}
\begin{figure*}[htbp]
    \centering
    \includegraphics[width=1.0\textwidth]{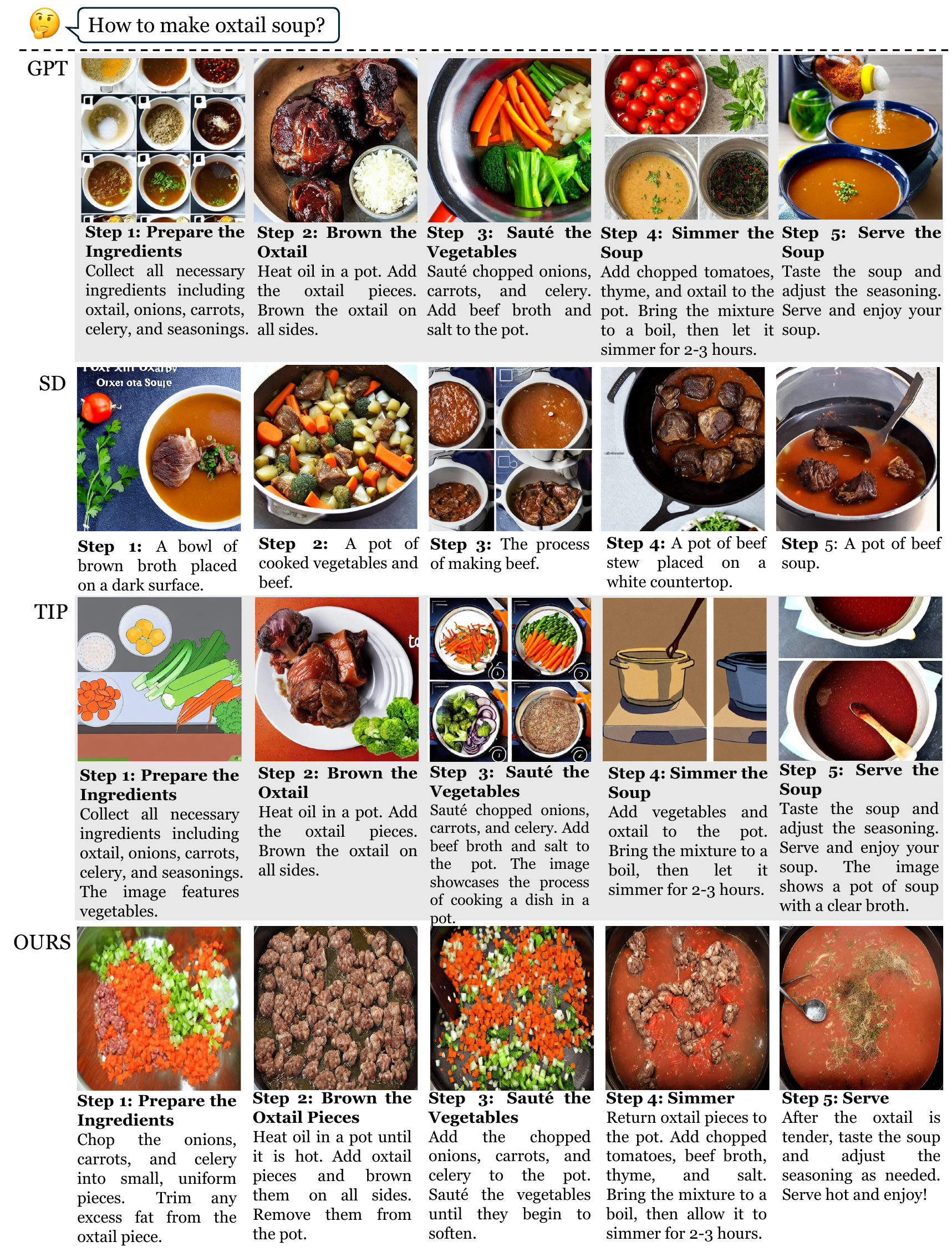}
    \caption{A qualitative comparison of plans output by all three baselines and our approach.}
    \label{fig:results-demo}
\end{figure*}

\vspace{-2mm}
\subsection{Main Results}
\label{section-results}
Table \ref{table:main-results} shows automatic measurements and LLM evaluation results with three different backbone models. Despite value differences across three sets of experiments, they exhibit the same trends in all three aspects. \textsc{Ours} approach obtains consistently higher scores on all automatic measurements of textual plans. In LLM evaluation, the coherence of textual plans generated by \textsc{Ours} approach ranks second to the vanilla baselines \textsc{Gemini}, \textsc{Gpt}, and \textsc{M\&L}. The potential cause is that generating the whole plan in one shot prevents semantic conflicts and temporal inconsistency. It is also noteworthy that all baselines and \textsc{Ours} approach obtain better scores in terms of textual plan evaluation, but obtain worse scores in terms of visual plan and text-image pair evaluation on wikiHow tasks. The observation is accountable provided different plan lengths and language styles of two data sources as we discussed in Section \ref{section-dataset} and Appendix \ref{sec:app-dataset}.

\vspace{-2mm}
\subsection{Visual Coherence Analyses}



\vspace{-1mm}
From Table \ref{table:main-results}, \textsc{Ours} approach obtains consistent improvement in terms of PPL score and I-I score by LLM judge. Although \textsc{Sd} baseline achieves the best PPL score in all cases, the underlying reason is that the image generation model is incapable of actual ``task planning''. When the planning instructions and task goal $\mathcal{G}$ are fed into it, it concentrates on the project or action mentioned in $\mathcal{G}$ and outputs images about the topic. As shown in Figure \ref{fig:results-demo}, the visual plan generated by \textsc{Sd} fails to showcase expected object transitions in temporal logic. Therefore, the translated textual ``steps'' are semantically similar, further leading to lower PPL score.

For other baselines, we observe that they tend to involve sharp transitions in the visual plan. Figure \ref{fig:results-demo} intensively shows such failures in maintaining visual coherence. For example, the visual plan generated by \textsc{Gpt} presents a cooking course where consecutive steps appear irrelevant. It is due to the ignorance of its image generation module about the last visual step. In contrast, the \textsc{Ours} approach addresses this challenge by predicting the next image conditioned on both the textual instruction and the last visual step. Consequently, the cooking course is smooth with natural object transitions.

\vspace{-2mm}
\subsection{Text-Image Consistency Analyses}
\vspace{-1mm}

\textsc{Ours} approach outperforms all baselines in terms of CLIP score and T-I score by LLM judge. \textsc{Ours} approach ensures high consistency between two modalities with two passes that first use the textual draft to instruct image generation and then refine the draft with extracted information from the generated image.
As presented in Figure \ref{fig:results-demo}, the visual step in \textsc{Tip}'s plan digresses from the task goal and its corresponding textual instruction. It is hard to observe ``oxtail'' from step 3 on. The cause is that without awareness of the previous steps, the image generation module is easily attracted by the most mentioned objects in the text. On the other hand, \textsc{Ours} approach counteracts the textual noise with visual information from the last step. 

\vspace{-2mm}
\subsection{Human Evaluation}
\vspace{-1mm}
To complement our automatic and LLM evaluation, we conduct a human evaluation by recruiting 3 annotators. We randomly select 50 tasks as a validation dataset. Its category distribution is consistent with the overall dataset. 
We choose the strong baseline \textsc{Gpt} as one candidate and \textsc{Ours} with the same backbone as the other. We show the plans generated by these two models in random order, and the annotators are asked to make comparative annotations (win/tie/lose) between the generated plans in three aspects: textual plan quality, visual plan coherence, and text-image alignment. Please check Appendix \ref{app-human-eval} for more evaluation details.

Table \ref{table:human_eval} shows the human evaluation results. It verifies the effectiveness of \textsc{Ours} approach in all 3 dimensions and is generally in line with automatic and LLM evaluations. Compared with the baseline \textsc{Gpt}, \textsc{Ours} approach obtains slight improvements in textual plan quality. Regarding visual plan coherence and text-image alignment, \textsc{Ours} approach exhibits significant superiority.

\begin{table}[t!]
\small
\begin{tabular}{lll}
\toprule
\multicolumn{1}{c}{\multirow{2}{*}{\raisebox{-.5\height}{Eval.}}} & \multicolumn{1}{c}{\textsc{Ours} v.s. \textsc{Gpt-4o}} & \multicolumn{1}{c}{\multirow{2}{*}{\raisebox{-.5\height}{$\kappa$}}} \\ \cmidrule(lr){2-2}
\multicolumn{1}{c}{}                            & \multicolumn{1}{c}{\colorbox{lightblue}{Win}/\colorbox{lightgreen}{Tie}/\colorbox{lightyellow}{Lose}} & \multicolumn{1}{c}{}                       \\ \midrule
Text                            & \raisebox{-.3\height}{\includegraphics[width=0.65\linewidth]{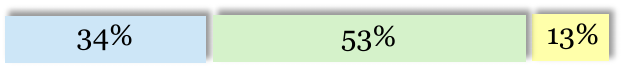}}                                 & 0.521                                           \\
Image                             & \raisebox{-.3\height}{\includegraphics[width=0.65\linewidth]{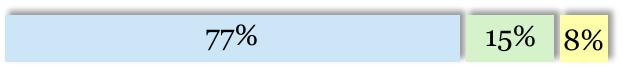}}                                & 0.604                                           \\
T-I    & \raisebox{-.3\height}{\includegraphics[width=0.66\linewidth]{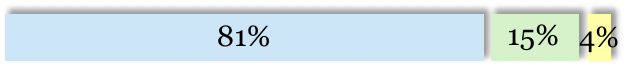}} & 0.699 \\
\bottomrule
\end{tabular}
\caption{Human evaluation results. The $\kappa$ scores demonstrate moderate to substantial inter-annotator agreement.}
\vspace{-3mm}
\label{table:human_eval}
\end{table}
\begin{table*}[t!]
\small
\begin{center}
\renewcommand{\arraystretch}{1.2}
\resizebox{\textwidth}{!}{
\begin{tabular}{clcccccccccccc}
\toprule
\multirow{2}{*}{\raisebox{-.5\height}{Dataset}}       & \multirow{2}{*}{\raisebox{-.5\height}{Method}} & \multicolumn{6}{c}{Automatic Evaluation}  & \multicolumn{6}{c}{LLM Evaluation}                        \\ \cmidrule(lr){3-8} \cmidrule(lr){9-14}
                               &                         & BertScore $\uparrow$ & R-1 $\uparrow$ & R-2 $\uparrow$ & R-L $\uparrow$ & CLIP $\uparrow$ & PPL $\downarrow$ & Corr. $\uparrow$ & Exec. $\uparrow$ & Coh. $\uparrow$ & Info. $\uparrow$ & T-I $\uparrow$ & I-I $\uparrow$ \\ \midrule
\multirow{4}{*}{Instructables} & \textsc{w Des}                   & \underline{0.841}      & \underline{0.295}   & \underline{0.083}   & \underline{0.279}   & 14.71          & 5.92      & \underline{4.63}       & 4.17      & \underline{4.34}      & \underline{4.58}      & \underline{1.76}      & \underline{2.39}\\
                               & \textsc{w Img}                   & 0.836      & 0.257   & 0.069   & 0.245   & \underline{16.48}          & \underline{5.90}       & 4.40      & \underline{4.24}      & 4.19      & 4.47      & 1.72      & 2.26\\
                               & \textsc{pPDDL-to-Nl}                   & 0.837      & 0.261   & 0.079   & 0.243   & 12.04          & 6.25       & 4.18      & 4.20      & 3.93      & 4.02      & 1.58      & 2.09\\
                               & \textsc{Ours}                    & \textbf{0.849}      & \textbf{0.337}   & \textbf{0.103}   & \textbf{0.315}   & \textbf{27.14}          &  \textbf{5.21}       & \textbf{4.93}      & \textbf{4.90}       & \textbf{4.93}      & \textbf{4.93}      & \textbf{2.47}      & \textbf{2.76}\\
                               \midrule
\multirow{4}{*}{wikiHow}       & \textsc{w Des}                   & 0.847      & \underline{0.309}   & \underline{0.092}   & \underline{0.290}    & \underline{12.04}   & 5.89      & \underline{4.60}       & 4.28      & 4.40      & \underline{4.84}      & \underline{1.85}      & \underline{2.34}\\
                               & \textsc{w Img}                   & \underline{0.849}      & 0.298   & 0.090   & 0.278    & 11.97   & \underline{5.76}       & 4.36      & \underline{4.31}      & \underline{4.48}      & 4.77      & 1.79      & 2.18\\
                               & \textsc{pPDDL-to-Nl}                   & 0.840      & 0.279   & 0.081   & 0.255   & 11.35          & 6.38       & 4.14      & 4.02      & 3.95      & 4.33      & 1.45      & 1.97\\
                               & \textsc{Ours}                    & \textbf{0.856}      & \textbf{0.332}   & \textbf{0.105}   & \textbf{0.309}    & \textbf{14.62}  & \textbf{5.30}       & \textbf{4.88}      & \textbf{4.91}      & \textbf{4.90}      & \textbf{4.94}      & \textbf{2.58}      & \textbf{2.68}\\
                               \bottomrule
\end{tabular}}
\end{center}
\vspace{-3mm}
\caption{Ablation study using GPT-4o backbone.}
\vspace{-3mm}
\label{table:ablation-gpt}
\end{table*}
\vspace{-2mm}
\subsection{Ablation Studies}
\vspace{-1mm}

To further examine the effects of our formatted visual information design, we implement three ablation studies with backbone \textsc{gpt-4o} including (1) \textsc{w Des} which replaces the formatted visual information $v_k$ in our framework with general image descriptions; (2) \textsc{w Img} which directly feeds the generated image $i_k$ and the draft $d_k$ into the multimodal text refinement module and obtains the output textual step $t_k$; and (3) \textsc{pPddl-to-Nl} which conversely asks the draft generator to provide formatted draft as image generation guidance and translates it into natural language afterward.
The experiment results are shown in Table \ref{table:ablation-gpt}.

\vspace{-2mm}
\paragraph{Formatted v.s. General Visual Information}

Aiming to study pPDDL's necessity, we edit the prompt in stage 3 to make the visual information extractor describe $i_k$ in natural language (NL). The results in Table \ref{table:ablation-gpt} indicate a performance drop induced by this modification. From its intermediate outputs, we find that without a pre-defined format restriction, the image-to-text model tends to generate lengthy, unstructured descriptions with much attention on minutia. They introduce noise to text refinement in stage 4. As our framework is autoregressive, the noise accumulates with time steps, harming both the textual and visual plans.

\vspace{-2.5mm}
\paragraph{Formatted v.s. Raw Visual Information}

Given that \textsc{gemini-1.5-flash} and \textsc{gpt-4o} are infused with visual knowledge, we seek to explore whether these models can directly use raw image $i_k$ to refine the draft $d_k$. In this case, stage 3 is skipped. Table \ref{table:ablation-gpt} demonstrates the failure of depending on their image interpretation capability to refine textual plans. Without an external visual information extractor, they are likely to roughly append their image interpretations to the input drafts, leading to worse plan coherence.

\vspace{-2mm}
\paragraph{NL-to-pPDDL v.s. pPDDL-to-NL}

To determine whether the pPDDL functions better than NL for image generation, we exchange the order of generating texts in two formats. In this ablation study, the draft generation module is prompted to generate a pPDDL step $d_k$ in stage 1. In stage 3, we ask the visual information extractor to describe $i_k$ in NL $v_k$. In stage 4, the text refinement is still based on both $d_k$ and $v_k$.

From Table \ref{table:ablation-gpt}, \textsc{Ours} approach outperforms the pPDDL-to-NL method. Inspecting the concrete generation results, we observe that the pPDDL-first design can not yield sufficient information for the image generation module in stage 2. The highly compact drafts make it difficult to generate visual counterparts that align with textual instructions. It further disables the visual information extractor from generating accurate image descriptions. The refined textual steps are often accordingly ambiguous. Therefore, the overall evaluation of this method is poor in all aspects.

\vspace{-2.5mm}
\subsection{Sensitivity to Task Complexity}
\vspace{-1mm}
\begin{figure}[t!]
    \centering
    \vspace{-5mm}
    \includegraphics[width=0.52\textwidth]{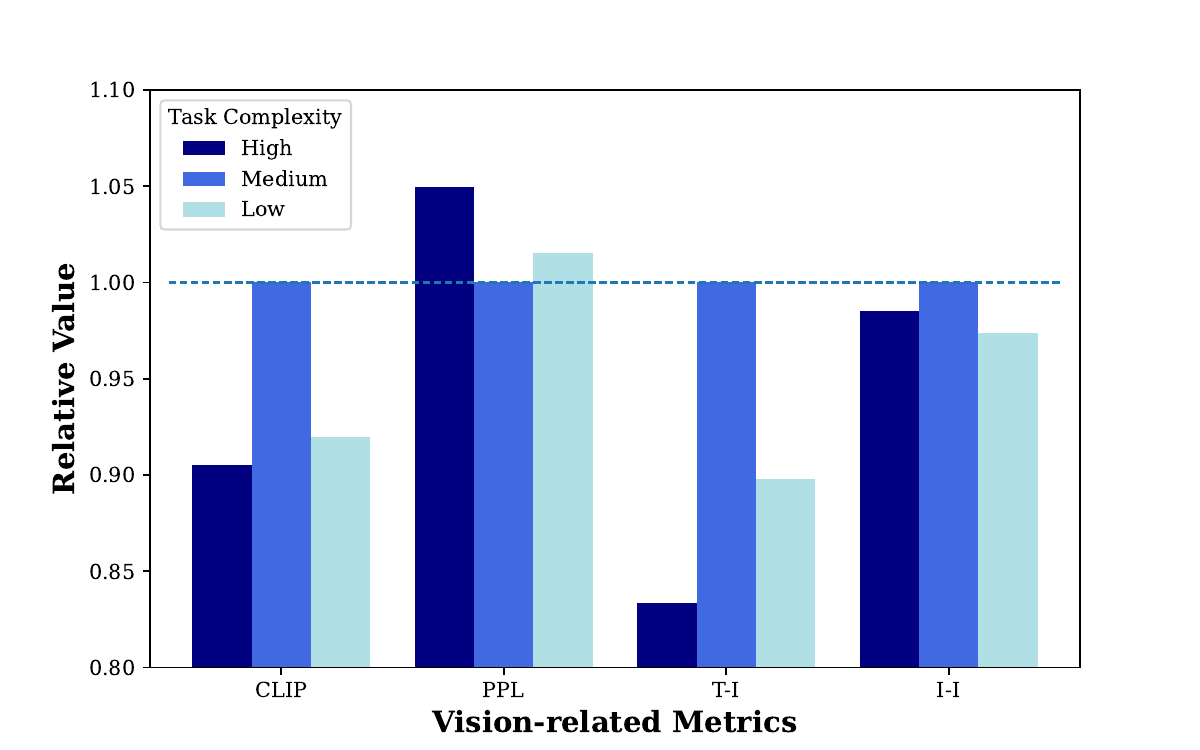}
    \vspace{-5mm}
    \caption{
   Visual step coherence and text-image alignment across different task complexity levels. All values are normalized relative to the results from medium-complexity tasks. For PPL score, the lower, the better.
    }
     \vspace{-3mm}
    \label{fig:complex-analysis}
\end{figure}

We observe that diverse task categories introduce different task complexities. For instance, ``Home \& Garden'' tasks are moderately challenging, and both textual and visual plans provide similar information. ``Hobbies \& Craft'' tasks are often complex and require delicate actions. In this case, the visual plan is more informative than the textual plan. ``Education \& Communications'' tasks are relatively abstract. The textual plan is better for such tasks since the visual plan always depicts concrete actions or scenarios. To study the effectiveness of our framework over varying task complexities, we sample 100 tasks and manually annotate them with complexity high, medium, or low.

Figure \ref{fig:complex-analysis} demonstrates our approach yields better visual plans when challenged with medium-complexity tasks. Involved actions in such plans are often concrete while not elaborate, enabling the image generation model to visualize them precisely. However, it struggles with high- and low-complexity tasks. Depicting abstract or delicate actions still exceeds the capacity of current models.



\vspace{-2.5mm}
\section{Conclusion}
\vspace{-2mm}

Our work studies an underexplored problem of text-image plan generation. We identify two main challenges: ensuring visual coherence and text-image alignment, and propose a novel framework to address them accordingly. For evaluation, we collect a dataset of daily tasks covering diverse domains and task complexities. Substantial experiment results demonstrate the effectiveness of our approach on a range of various backbone models, especially in terms of the two challenges we aim to address.
\section*{Limitations}

While our approach exhibits promising results in improving text-image plan generation, it is noteworthy that our work has several limitations.

First, given that LLMs are trained on vast amounts of data, data leakage is inevitable. This inherent characteristic potentially contributes to their strong performance in textual plan generation, as similar patterns may exist in their training data. Second, the quality of visual plans generated by fine-tuned InstructPix2Pix suggests room for improvement. Although it is capable of maintaining visual coherence when the scenario and object have minor transitions, we still observe unexpected incoherence when the textual instructions indicate significant workspace change. Last, the measurement of visual coherence through text-based metrics is indirect. The textual descriptions converted from images may not fully capture the nuanced visual relationships and coherence patterns that exist in the original images, potentially affecting the validity of our evaluation.

These limitations shed light on potential directions for future work, including the exploration of image editing models that better fit into the planning context, the development of visual coherence metrics that function in the image space, etc.
\section*{Ethics Statement}
Instructables and wikiHow are two public online platforms licensed under CC BY-NC-SA 3.0 and CC BY-NC-SA 4.0, respectively. Our data collection process complies with their licensing terms, as both licenses permit academic use with proper attribution. Our work primarily focuses on generating text-image plans for daily tasks. Therefore, we exclude potentially inappropriate content in the process of data collection and manually inspect data quality. 
To prevent privacy leakage, we anonymize any personal information of the plan authors. 
Our human evaluation is conducted by three graduate students who are co-authors of this paper. They participate in the evaluation process as part of their research contribution and are acknowledged through co-authorship.

Last, our approach relies on LLMs, which may produce inconsistent or biased results. While designed for daily task planning, this approach could potentially be misused for malicious intents. Future work should investigate these risks and develop additional safeguards.



\section*{Acknowledgments}

This work was supported by NSF CAREER Award IIS-2338418. 

\bibliography{acl_latex}

\appendix




\section{Dataset}
\label{sec:app-dataset}
\subsection{Data Sources}
\begin{table}[ht]
\centering
\small
\begin{tabular}{l|cc}
\toprule
                          & Instructables & wikiHow \\ \midrule
\# of tasks               & 100           & 1,000   \\
\# of task categories     & 1             & 10      \\
Avg. \# of steps per task & 7.20          & 33.30   \\
Avg. \# of words per step & 9.76          & 45.84   \\ \bottomrule
\end{tabular}
\caption{Statistics of data collected from two sources.}
\label{table:data_statistics}
\end{table}
\begin{figure*}[htbp]
    \centering
    \includegraphics[width=1.0\textwidth]{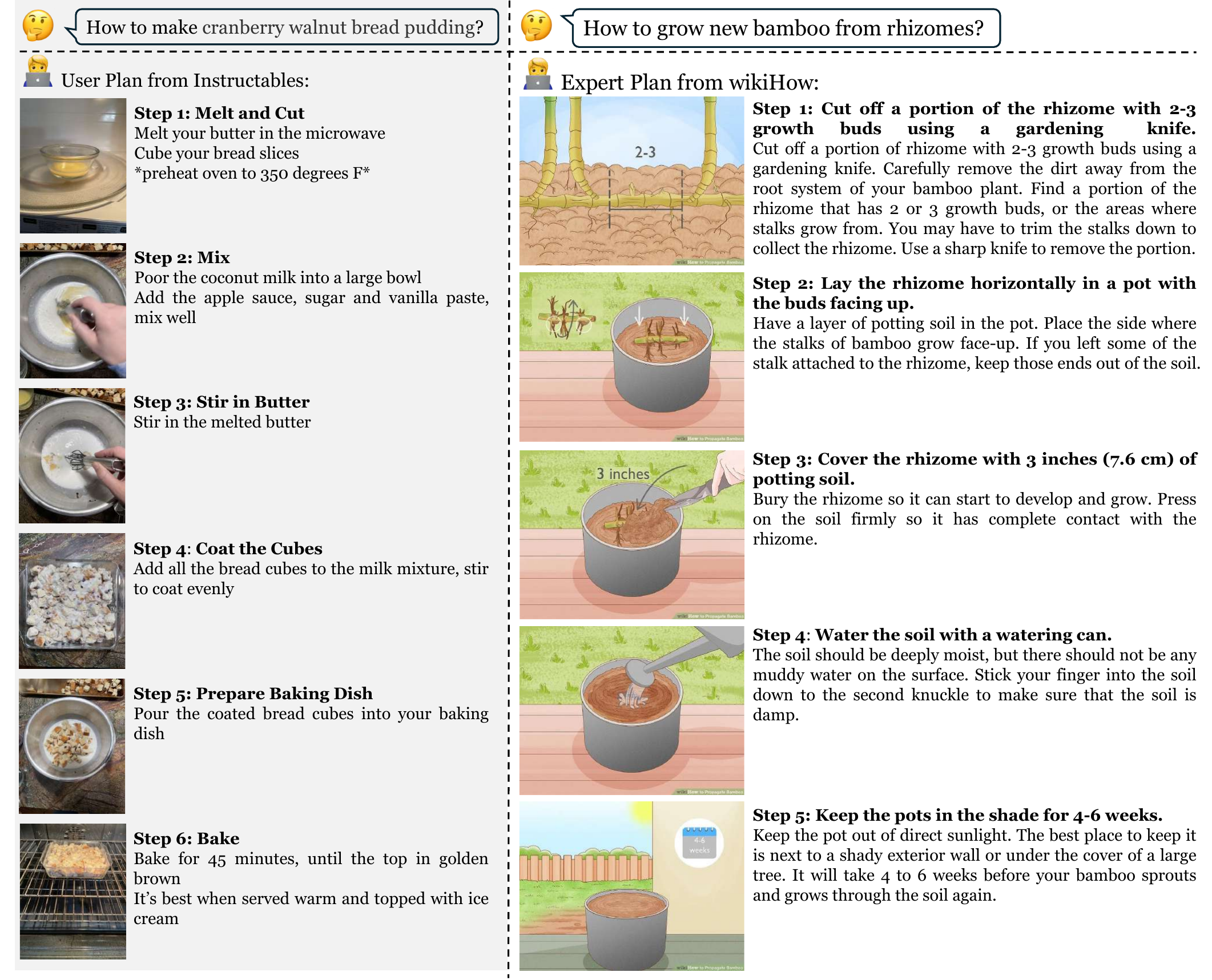}
    \caption{Two example plans sampled from Instructables data (left) and wikiHow data (right) respectively.}
    \label{fig:data-examples}
\end{figure*}

To provide more insight into the two sources of our dataset, we further present both quantitative and qualitative comparisons between their plans. As discussed in Section \ref{section-dataset}, they exhibit differences in plan lengths and language styles. As Table \ref{table:data_statistics} reports, the average length of wikiHow plans exceeds Instructable plans in both the number of steps per plan and the number of words per step.

Figure~\ref{fig:data-examples} demonstrates an intuitive distinction between two data sources. The left-hand side plan from Instructables is brief in textual descriptions while the right-hand side plan from wikiHow is elaborate with details including execution tips and potential outcomes. The accompanying images are also different in style.

\subsection{Task Complexity Classification}

Our task complexity classification is based on the relative informativeness of textual versus visual planning modalities, reflecting the inherent challenges of different procedural domains.

\begin{figure*}[htbp]
    \centering
    \includegraphics[width=1.0\textwidth]{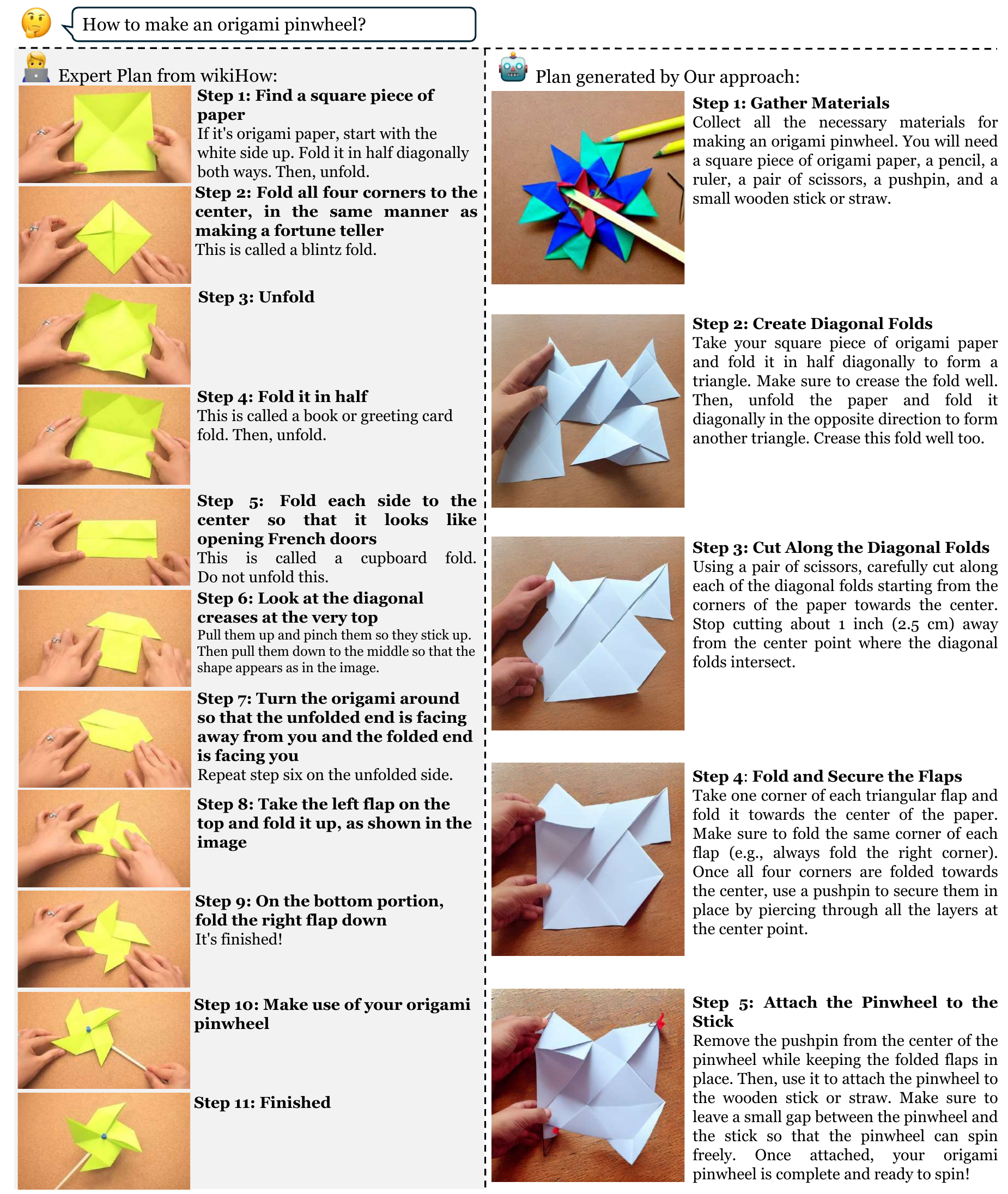}
    \caption{Expert plan from wikiHow vs. plan generated by \textsc{Ours} approach for a high-complexity task \textit{How to make an origami pinwheel}.}
    \label{fig:high-example}
\end{figure*}

\paragraph{High complexity tasks} This type of task typically involves intricate procedures that require delicate actions, precise object transitions, or spatial reasoning. For such tasks, visual plans significantly outperform textual descriptions in conveying critical procedural information. Figure \ref{fig:high-example} shows a task``make an origami pinwheel'', where visual demonstration of folding sequences and spatial relationships is essential for successful execution. In this case, \textsc{Ours} approach faces constraints imposed by current text-image-to-image model capabilities. Despite the fine-tuning, these models struggle to accurately depict subtle hand movements, precise folding sequences, and fine-grained spatial transformations essential for such a complex crafting task.

\begin{figure*}[htbp]
    \centering
    \includegraphics[width=1.0\textwidth]{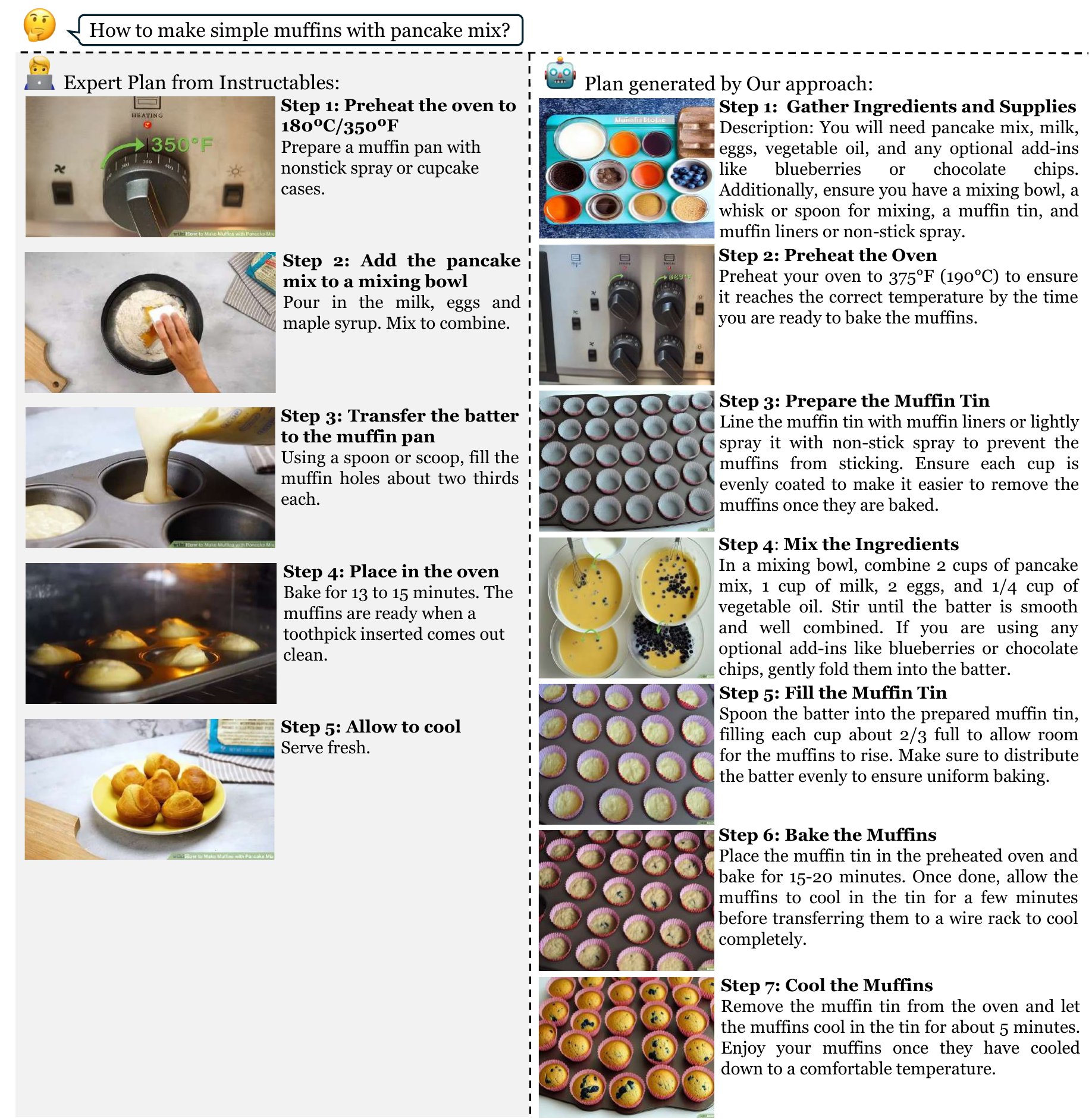}
    \caption{Expert plan from Instructables vs. plan generated by \textsc{Ours} approach for a medium-complexity task \textit{How to make simple muffins with pancake mix}.}
    \label{fig:med-example}
\end{figure*}

\paragraph{Medium complexity tasks} This type of task often involves moderately challenging procedures where textual and visual plans provide comparable information. Figure \ref{fig:med-example} shows a task ``make simple muffins with pancake mix'', where both modalities effectively convey the sequential steps and key procedural elements. \textsc{Ours} framework achieves optimal performance in such tasks where the procedural knowledge does not demand extreme precision or highly abstract concepts.

\begin{figure*}[htbp]
    \centering
    \includegraphics[width=1.0\textwidth]{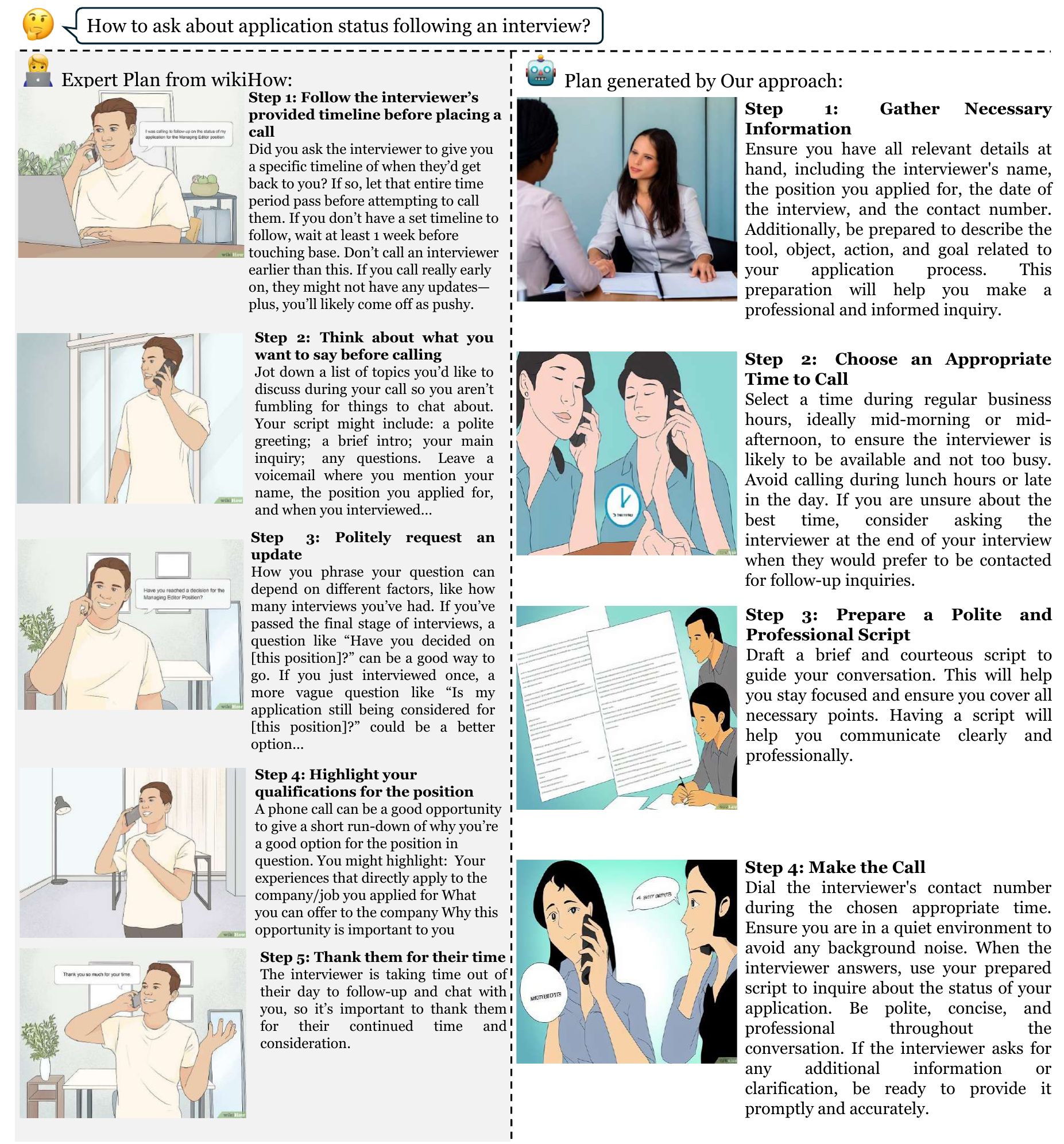}
    \caption{Expert plan from wikiHow vs. plan generated by \textsc{Ours} approach for a low-complexity task \textit{How to ask about application status following an interview}.}
    \label{fig:low-example}
\end{figure*}

\paragraph{Low complexity tasks} This type of task encompasses relatively abstract or conceptual procedures where textual plans are more informative than visual representations. Figure \ref{fig:low-example} shows a task ``ask about application status following an interview'', where the procedural knowledge is primarily communicative and contextual rather than visually demonstrable. In this case, the underlying textual instructions are inherently abstract and focus on communicative strategies, timing, and contextual considerations. Therefore, the visual plan generated by \textsc{Ours} approach becomes less descriptive and struggles to represent non-concrete actions, resulting in generic and superficial representations.


\section{Experiment Settings}
\label{app-exp-settings}

\subsection{Prompt Design}
\begin{table*}[ht]
\small
\begin{center}
\renewcommand{\arraystretch}{1.2}
\resizebox{\textwidth}{!}{
\begin{tabular}{ll}
\toprule
Stage  & Prompt                        \\ \midrule \midrule
Draft Generation       & \makecell[l]{You are a helpful planning assistant. Let's break down how to achieve the following goal step by step: \\
GOAL: $\mathcal{G}$ \\ PREVIOUS STEPS: $t_{1...k-1}$ \\ What is the next specific, actionable step toward achieving this goal? \\ Please format your response as: \\ STEP [$k$]: [Step Title] [step descriptions] \\ After providing the step, indicate if the goal is achieved [YES/NO].}        \\ \midrule
Image Generation       & A clear, detailed photograph showing $d_k$, high quality, realistic with natural lighting        \\ \midrule
\makecell[l]{Visual Info\\Extraction }      & \makecell[l]{The provided image shows a step to $\mathcal{G}$. Please analyze it and extract the following information:\\ 1. Objects: List salient objects in the image\\2. Tools: Identify any tools, equipment, or instruments being used\\3. Actions: Describe the specific actions being performed\\4. Goal: Based on the visible actions and context, what appears to be the intended goal?\\Please format your response as:\\OBJECTS: [object list]\\TOOLS: [tool list]\\ACTIONS: [action list]\\GOAL: [state the apparent goal]}        \\ \midrule
Text Refinement       & \makecell[l]{You are a helpful planning assistant. Let's improve a step to $\mathcal{G}$ with visual information.\\Original Step: $d_k$\\Visual Information Extracted: $v_k$\\The improved step should:\\1. Be more specific about the objects and tools involved\\2. Provide clearer action descriptions\\3. Maintain alignment with the overall goal\\Please format your response as: [improved step descriptions]}        \\ 
                               
\bottomrule
\end{tabular}}
\end{center}
\caption{Prompt templates we use in experiments with backbone model Gemini-1.5-flash.}
\label{table:prompt-exp}
\end{table*}
In this section, we present our prompt design as a reference. Table \ref{table:prompt-exp} shows the prompts we use in all four stages of our framework with backbone Gemini-1.5-flash. We make only trivial adjustments for the other two backbone models and omit them for brevity and clarity.

\subsection{Computational Costs}
Our framework incurs computational costs from two primary sources: LLM inference during testing and one-time InstructPix2Pix fine-tuning. During inference, each task requires approximately 21 LLM calls distributed across plan drafting ($\sim$7 calls), visual information extraction ($\sim$7 calls), and iterative plan refinement ($\sim$7 calls), plus inference from our fine-tuned InstructPix2Pix model for generating images at each planning step. The InstructPix2Pix fine-tuning represents a one-time computational investment, requiring 4 hours on 4x NVIDIA A100 80GB GPUs using 7,500 text-image-to-image pairs.

\section{Evaluation}
Our comprehensive evaluation strategy employs three complementary assessment dimensions. This multi-faceted approach provides a complete picture of our framework's performance, addressing the inherent limitations of individual evaluation methods while leveraging their respective strengths. The convergent evidence across all three dimensions strengthens the validity of our findings.
\subsection{LLM Evaluation}
\begin{table*}[ht]
\small
\begin{center}
\renewcommand{\arraystretch}{1.2}
\resizebox{\textwidth}{!}{
\begin{tabular}{ll}
\toprule
Evaluation  & Prompt                        \\ \midrule \midrule
Textual Plan       & \makecell[l]{You are a helpful evaluation assistant. Please assess the following plan to $\mathcal{G}$ against the provided reference plan using these four criteria:\\1. Correctness: Does the plan contain all necessary steps that align with the reference? This involves checking if the steps are complete.\\2. Executability: How practical and actionable are the steps? This involves checking if they can be implemented in a real-world setting.\\3. Coherence: Are all steps logically connected to each other? This involves checking if there are temporal conflicts or redundancy.\\4. Informativeness: Does the plan provide sufficient detail? This involves checking if it provides enough information to understand the plan.\\Grading scale: 1-Poor 2-Fair 3-Good 4-Very Good 5-Excellent\\Reference Plan: [reference plan $\mathcal{R}$]\\Plan to Evaluate: [evaluated plan $\mathcal{P}$]\\Please provide a numeric score and a brief justification for each criterion.}        \\ \midrule
Visual Plan       & \makecell[l]{You are a helpful evaluation assistant. Please assess how well Image 2 continues from Image 1 considering the provided step description.\\Step description: $t_k$\\Grading Scale:\\1-Poor: images appear unrelated or contradictory\\2-Fair: slight logical connection but major inconsistencies\\3-Good: clear connection but some inconsistencies\\4-Very good: strong connection with minor inconsistencies\\5-Excellent: perfect logical progression\\Please provide a numeric score and a brief justification.}        \\ \midrule
\makecell[l]{Text-Image Alignment}      & \makecell[l]{You are a helpful evaluation assistant. Please evaluate how well the provided image aligns with the given step description.\\Step description: $t_k$\\Grading Scale:\\1-Poor: image appears unrelated to the step\\2-Fair: image partially reflects the step but has major mismatches\\3-Good: image mostly reflects the step with some mismatches\\4-Very good-mage clearly reflects the step with minor mismatches\\5-Excellent: image perfectly represents the step\\Please provide a numeric score and a brief justification.}        \\
                               
\bottomrule
\end{tabular}}
\end{center}
\caption{Prompt templates we use for LLM evaluation.}
\label{table:prompt-eval}
\end{table*}
Table \ref{table:prompt-eval} presents prompts we use for LLM evaluation. 
Regarding textual plan evaluation, we define four aspects with inspirations from \citet{huang2022languagemodelszeroshotplanners}'s metrics design: correctness, executability, coherence, and informativeness. Their definitions are shown in Table \ref{table:prompt-eval}.
For visual plan evaluation, the evaluator Claude-3.5-Sonnet receives two images $(i_{k-1},i_k)$ in addition to the textual prompt. We prompt the MLLM to measure if $i_k$ logically follows $i_{k-1}$ considering the potential effect rendered by the step description $t_k$. For text-image alignment evaluation, the evaluator receives an image $i_k$. We prompt the MLLM to measure if $i_k$ is semantically aligned with $t_k$. The grading criteria are elaborated in Table \ref{table:prompt-eval}.

\subsection{Human Evaluation}
\label{app-human-eval}

\begin{figure*}[htbp]
    \centering
    \includegraphics[width=1.0\textwidth]{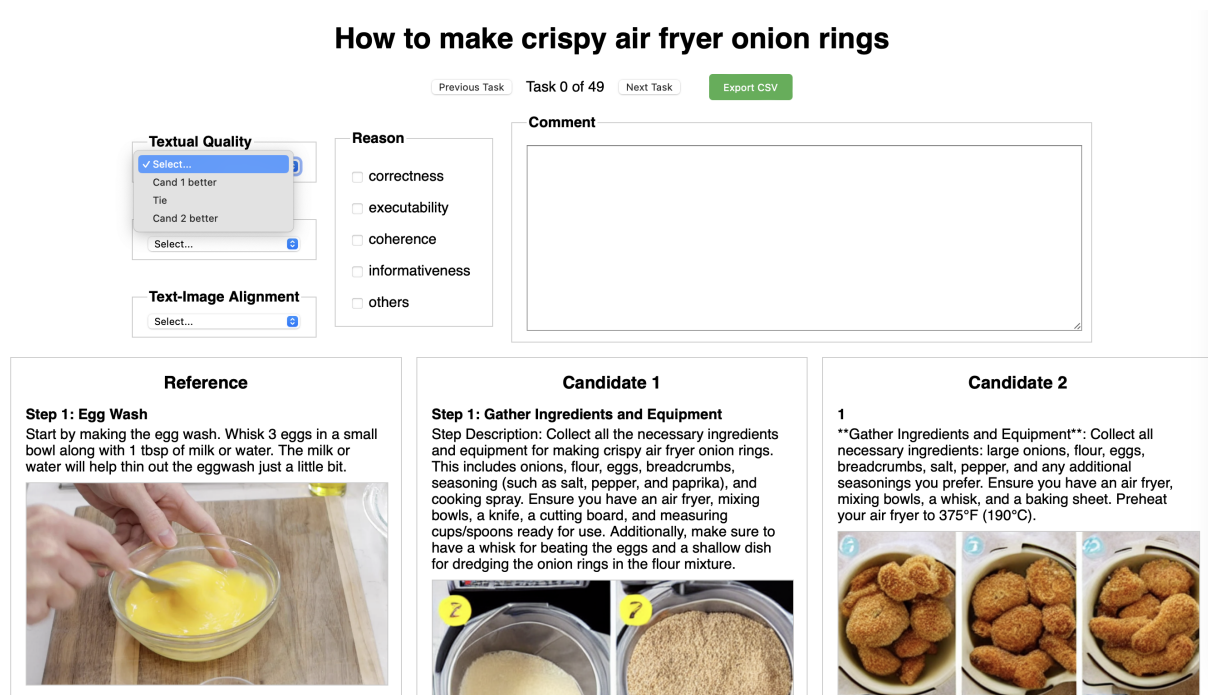}
    \caption{User interface of our designed annotation tool.}
    \label{fig:ui}
\end{figure*}
To facilitate human evaluation, we design an annotation tool as shown in Figure \ref{fig:ui}. In addition to two candidate plans, we also provide the reference article from the original data source (Instructables and wikiHow) in case annotators are unfamiliar with the task field. For every evaluation aspect, we provide three options: Candidate 1 better, Tie, and Candidate 2 better. To align with LLM evaluation, we list potential reasons annotators do not choose ``Tie'' for ``Textual Quality''. For example, if the annotator chooses ``Candidate 1 better'' when they evaluate textual plans, and they find its superiority over Candidate 2 is mainly in ``coherence'' and ``informativeness'', they are required to tick these two reasons. Furthermore, there is a text field for annotators to mark down their observations regarding any plan quality issues.

\end{document}